\documentclass[10pt, a4paper]{article}
\usepackage{lrec2000}
\usepackage{qtree,alltt,epsfig,url,xy}
\xyoption{all}
\def\myurl#1{{\small[\url{#1}]}}
\newenvironment{sv}{\scriptsize\begin{alltt}}{\end{alltt}\normalsize}

\title{An Integrated Framework for Treebanks and Multilayer Annotations}

\name{Scott Cotton and Steven Bird}

\address{
  Department of Computer and Information Science\\
  200 South 33rd Street, University of Pennsylvania\\
  Philadelphia, PA 19104-6389, USA\\
  {\small\tt cotton@linc.cis.upenn.edu, sb@ldc.upenn.edu}
}

\abstract{
Treebank formats and associated software tools are proliferating rapidly,
with little consideration for interoperability.  We survey a wide variety
of treebank structures and operations, and show how they can be mapped onto
the annotation graph model, and leading to an integrated framework
encompassing tree and non-tree annotations alike.  This development opens up new
possibilities for managing and exploiting multilayer annotations.
}

\begin{document}

\maketitleabstract

\section{Introduction}
\label{sec:intro}

There is a proliferation of treebanks, each with its own format and
software tools.  Examples include the Penn Treebank, the Prague Dependency
Treebank, the Susanne Corpus, and treebanks of German, Spanish, Portuguese,
French, Italian, Turkish, Polish, Bulgarian, Old English, and the recent
development of Korean, Arabic and Chinese treebanks.  Each treebank is
associated with tools for annotation, search, and rendering.  Despite the
obvious benefits of interoperability, the tools associated with any given
treebank rarely escape the confines of its own project.  Moreover,
treebanks both require and invite multilayer capabilities.  Parsers depend
on tokenizers, taggers, and morphological analyzers.  Layers of annotation
such as sense tagging and named entity tagging are built on top of
treebanks.  Disfluency tagging, as combined with treebanking in
switchboard, adds another layer of indirection between parses and the
surface string.  In short, necessity dictates the integration of treebanks
into a general multilayer annotation system, coupled with the development
of a logical model and corresponding API which address the linguistic
demands of treebanking.

Linguistically, the development of such a framework leads to some interesting
challenges.  Grammars and theories of syntax yield structures which stretch the
simplistic notion of trees over surface strings (such as empty constituents,
encoding of deep syntactic structure, pure dependency structures, etc).  As
advances in information extraction and language understanding bridge syntax and
semantics, syntactic trees are growing various forms of semantic annotation.  A
case in point is the English Propbank, in which sentences are annotated with
many fine grained semantic relations (or propositions) whose arguments in turn
point to relevant syntactic substructures such as individual nodes or trace
chains.  The design and development of a system which aptly addresses these
issues is certainly nontrivial.

In this paper we examine convetional phrase structure trees, dependency
trees, and semantic trees.  In each of these categories, we first survey
the data formats and editing operations, outline an abstract API for the
structural operations involved, and describe an implementation with
annotation graphs.

A variety of treebank formats and models are covered by the survey.
Sources of this variety are both linguistic and computational.  On the
linguistic side, languages may permit a greater or lesser degree of
word-order freedom.  In some cases, the conventional tree representation
requires crossing branches.  This happens to a limited extent in English,
with phenomena such as adverbials and extraposition.  However it is
pervasive in languages having rich case-marking systems such as Czech.
Treebanks for these languages typically use a dependency representation
instead of the conventional tree representation.  On the computational
side, projects may have different prior commitments to file formats.  The
file format may simply be derived from the original Penn Treebank format,
or be a novel plain text format, or be one of a variety of possible XML
representations.  To a considerable extent, these formats are
inter-translatable.  Another source of variation is the kind of information
which is annotated, and the survey includes some recent work on semantic
annotation and predicate-argument tagging.

After reviewing a diverse set of treebank projects, we consider the kinds
of tree-manipulation operations they require, leading to an inventory of
elementary tree operations.  These operations may be composed with each
other to perform complex tree manipulations.  Next, we show how the operations
can be implemented in the annotation graph model \cite{BirdLiberman01}.
This mapping has an important consequence for multilayer annotations,
for now treebanks can co-exist with a variety of other annotation
types, such as prosodic and discourse level annotations.  With all the
annotations expressed in the same data model, it becomes a straightforward
matter to investigate the relationships between the various linguistic
levels.  Modeling the interaction between linguistic levels is a central
concern both for the study of human communicative interaction, and for
the construction of naturalistic spoken language dialogue systems.

This inventory of elementary tree operations leads to a new application
programming interface for treebanking, built on top of the existing
annotation graph API which is implemented in the Annotation Graph
Toolkit \cite{MaedaBird02}.  This implementation work is ongoing, and
will be released under an open source license with AGTK.

\section{Conventional Syntax Trees}
\label{sec:conventional}
\subsection{Survey}
\label{sec:consurv}

The Penn Treebank was the first syntactically annotated corpus, and
consists of one million words of manually parsed text from the
Wall Street Journal \cite{Marcus93}.  An example of the Treebank format
is shown below.

\begin{sv}
((S (NP-SBJ-1
      (NP Yields)
      (PP on
        (NP money-market mutual funds)))
    (VP continued
      (S (NP-SBJ *-1)
         (VP to
           (VP slide)))
      ,
      (PP-LOC amid
        (NP signs
          (SBAR that
            (S (NP-SBJ portfolio managers)
              (VP expect
                (NP (NP further declines)
                  (PP-LOC in
                    (NP interest rates)))))))))
.))
\end{sv}

The empty constituents, called {\em traces}, represent
various forms of syntactic movement that serve to normalize the underlying
grammar.  In this example there is a trace \texttt{*-1},
immediately preceding the infinitive {\em to slide}.  This node is an empty
consituent and refers to the phrase {\em Yields on money-market mutual funds},
as is indicated by the fact that both nodes share the \texttt{-1} label.  The
movement of this nominal phrase to the nominal position in the infinitival
\texttt{S} clause normalizes this clause so that its constituents are
\texttt{NP} followed by \texttt{VP}.   In the Penn Treebank, traces are also
used to indicate WH and other pronominal movement.  Full details can be found
in the annotation guidelines.

The data in the Penn Treebank were created with an Emacs mode called {\em
parser-mode}.  The tool parses files of bracketted text in various stages of
the corpus development, starting with output from the automatic parser.
Editing operations add function tags, relabel, coindex, insert, and
delete constituents, and relocate subtrees from constituent to constituent.
Each of these operations is associated with a handful of constraints, such as
preservation of the surface string.  Notably, this constraint on the tree
editing operations leads to a set of tree editing functions for the user closed
under the following structural manipulations:  promotion of a leftmost or
rightmost constituent, insertion of a constituent, insertion/deletion of empty
constituents, and the movement of a constituent to its right (left,
respectively) sibling's leftmost (rightmost, respectively) child position.

In the rest of this section we consider various extensions to the
Penn Treebank format.

The Switchboard corpus of conversational speech \cite{Godfrey92}
was later enriched with information about breath groups and
disfluencies \cite{Taylor95}.  This new information is simple
enough on its own, e.g.:

\begin{sv}
B.22:   Yeah, / no one seems to be adopting it. /
  Metric system, [ no one's very, + {F uh, } no one wants ]
  it at all seems like. / 
\end{sv}

However, the disfluency information was also superimposed on
the syntactic trees, resulting in extremely complex structures
such as the following:

\begin{sv}
((S (NP-SBJ-1 no one)
    (VP seems
        (S (NP-SBJ *-1) 
           (VP to (VP be (VP adopting (NP it)))))) . 
               E_S))
((S (NP-TPC Metric system) ,
    (S-TPC-1 (EDITED (RM [)
                     (S (NP-SBJ no one)
                        (VP 's (ADJP-PRD-UNF very))) ,
                     (IP +)) (INTJ uh) ,
             (NP-SBJ no one)
             (VP wants (RS ]) (NP it) (ADVP at all)))
    (NP-SBJ *)
    (VP seems (SBAR like (S *T*-1))) . E_S))
\end{sv}

This format demonstrates the acute problem that arises when we
attempt to force one linguistic structure into a format that
was designed for representing a completely different kind of structure.

A more conservative extension of the Penn Treebank format is the
UAM Spanish Treebank \cite{Moreno00}
In this format, the treebank node labels have a record structure:

\begin{sv}
(S
  (NP SUBJ ID-1 SG P3
    (ART "<El>" "el" DEF MASC SG)
    (N "<Gobierno>" "Gobierno" SG P3))
  (VP TENSED PRES IND SG P3
    (V "<quiere>" "querer" TENSED PRES IND SG P3)
    (CL INFINITIVE OBJ1
      (NP * SUBJ REF-1)
      (VP UNTENSED INFINITE
        (V "<subir>" "subir" UNTENSED INFINITE)
        (NP OBJ1
          (ART "<los>" "el" INDEF MASC PL)
          (N "<impuestos>" "impuesto" MASC PL))))))
\end{sv}

Emacs is used for creating the structures, and a tree display tool is used for
verification.  Various other tools check for well formedness (e.g. of the node
attributes and grammatical structures).

Other treebanks use the same conventional nested structure, but
with a different syntax.  For example, consider the following
fragment from the Portuguese Treebank
\myurl{http://cgi.portugues.mct.pt/treebank/}.

\begin{sv}
<s>
SOURCE: CETEMPúblico n=1 sec=clt sem=92b
C1-2  O 7 e Meio é um ex-libris da noite algarvia.
A1
STA:fcl
SUBJ:np
=>N:art(M S)	O
=H:prop(M S)	7_e_Meio
P:v-fin(PR 3S IND)	é
SC:np
=>N:art(<arti> M S)	um
=H:n(M P)	ex-libris
=N<:pp
==H:prp(<sam->)	de
==P<:np
===>N:art(<-sam> F S)	a
===H:n(F S)	noite
===N<:adj(F S)	algarvia
.
</s>
\end{sv}

Emacs macros are used to edit the data, with operations for insertion and
deletion of nodes as well as increasing and decreasing the depth of the nodes
in the tree.  Some tree structural constraints are enforced: whenever a node's
depth is increased, so are all of its constituents, and all nodes must have a
label.

Finally, XML is now being used to represent treebanks.  The simplest and
most direct way to do this is to use element nesting to represent
hierarchy.  An example of this use of XML is provided by the
French Treebank \cite{Abeille00}, and we show a translation below.
\myurl{http://treebank.linguist.jussieu.fr/}.

%
%
\begin{sv}
<S>
  <NP>The proportion:NC
    <PP>of:P students:NC</PP>
  </NP>
  <PP>compared to:P
    <NP>the population:NC
      <PP>of:P
        <NP>our:D country:NC</NP>
      </PP>
    </NP>
  </PP>
  <PONCT>,:PONCT</PONCT>

[rest of sentence elided]

</S>
\end{sv}
It is notable that the part of speech labels are structured by convention in
the embedded text rather than by using XML markup.

\subsection{API}
\label{sec:conapi}

Many conventional tree operations, such as adding, moving or deleting
a subtree, also modify the sequence of terminals (or leaves).  In syntactic
annotation, this sequence is usually fixed, since it is an external artefact
which is not subject to editing.  Therefore, we need to provide a complete
inventory of tree operations which preserve the terminal string.

Many treebanking projects incorporate a preprocessing phase, which may
create some low-level constituents (such as noun phrase chunking) or may
create an entire parse of the sentence.  Therefore, the inventory of tree
operations must be capable of reorganizing the structure of an existing
tree, not just building a tree from scratch.

In this section we define an inventory of elementary tree operations which
preserve the terminal string and which is sufficiently expressive to
permit any well-formed phrase-structure tree to be built over the
terminal string, beginning either from an unparsed string or from a
previously parsed string.  The inventory is inspired by the various operations
that are provided by existing tree annotation tools.  We consider
only those operations which modify the structure of a tree (as opposed
to the operations for modifying node labels).

Each operation requires a tree $t$ along with a selected node $n$.
We write $t_n$ for the tree $t$ oriented at node $n$.

\begin{description}
\item[move down] $m_{\downarrow}(t_n)$
This creates a new node $\bar{n}$ in the position formerly occupied by $n$,
and makes $n$ the sole child of $\bar{n}$.  The new node $\bar{n}$ is an unlabeled
non-terminal symbol.  For example, under this operation, the tree on the
left becomes the tree on the right:

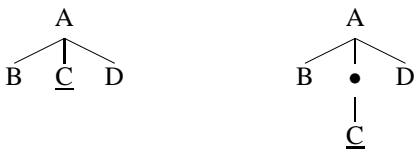
\begin{tabular}{ll}
\Tree [.A B \underline{C} D ]
&
\Tree [.A B [.{$\bullet$} \underline{C} ] D ]
\end{tabular}

\item[move up] $m_{\uparrow}(t_n)$
This applies only if $n$ has no siblings, deleting $\bar{n}$, the parent of
$n$.  Node $n$ now occupies the former position of $\bar{n}$.

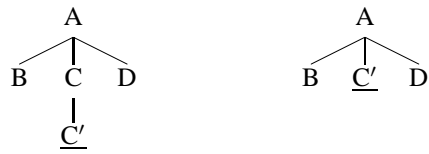
\begin{tabular}{ll}
\Tree [.A B [.C \underline{C$'$} ] D ]
&
\Tree [.A B \underline{C$'$} D ]
\end{tabular}

\item[promote right] $m_{\nearrow}(t_n)$
This applies only if $n$ has at least one sibling, but no siblings to its
right.  Node $n$ is moved up to the position immediately to the right of
its parent $\bar{n}$.

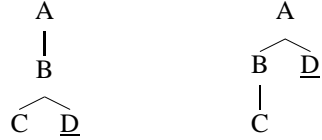
\begin{tabular}{ll}
\Tree [.A [.B C \underline{D} ] ]
&
\Tree [.A [.B C ] \underline{D} ]
\end{tabular}

\item[promote left] $m_{\nwarrow}(t_n)$
mirror image operation of $m_{\nearrow}(t_n)$.

\item[demote right] $m_{\searrow}(t_n)$
This applies only if $n$ has a sibling to the right $\overrightarrow{n}$,
and $\overrightarrow{n}$ is a non-terminal.  Node $n$ becomes the leftmost
child of $\overrightarrow{n}$.

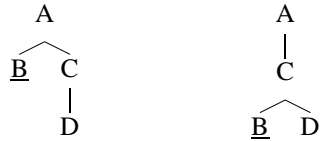
\begin{tabular}{ll}
\Tree [.A \underline{B} [.C D ] ]
&
\Tree [.A [.C \underline{B} D ] ]
\end{tabular}

\item[demote left] $m_{\swarrow}(t_n)$
mirror image operation of $m_{\searrow}(t_n)$
\end{description}

All operations preserve the orientation of the tree; the
selected node remains selected after the operation.  Observe
that all operations have inverses:
$m_{\downarrow\!\uparrow}(t_n) = t_n$;
$m_{\nearrow\!\!\!\swarrow}(t_n) = t_n$;
$m_{\nwarrow\!\!\!\searrow}(t_n) = t_n$.
All of these operations preserve the order of the terminal string,
and all are elementary as none can be expressed as a combination of
any others.

More complex operations can be built from these elementary operations.
For instance, in a particular user interface, it may be possible for a user
to select a set of contiguous terminals and and non-terminals, and group them under
a new non-terminal:

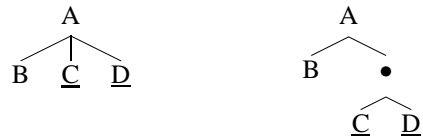
\begin{tabular}{ll}
\Tree [.A B \underline{C} \underline{D} ]
&
\Tree [.A B [.{$\bullet$} \underline{C} \underline{D} ] ]
\end{tabular}

\noindent
This can be done with a sequence of operations:
$m_{\downarrow}(t_C), m_{\swarrow}(t_D)$.
This is a generalized \textbf{move down} operation, for which
there is an corresponding generalized \textbf{move up}.

Note that there is another pair of elementary operations not discussed
above, that could be called trace-insertion and trace-deletion.  These
involve the creation/deletion of a zero width element in the terminal
sequence (or equivalently, of a ``non-terminal'' which dominates no
terminal).

\subsection{Implementation}
\label{sec:conimpl}

Bird and Liberman have developed a model for expressing the logical
structure of linguistic annotations, and have demonstrated that it can
encode a great variety of existing annotation types \cite{BirdLiberman01}.
An annotation graph is a directed acyclic graph where edges are labeled
with fielded records, and nodes are (optionally) labeled with time offsets.
The model is implemented in the Annotation Graph Toolkit and used as
the basis for several annotation tools, including one for editing
conventional syntax trees \cite{MaedaBird02,BirdMaeda02}.

Annotation graphs can most easily be used to represent trees using the
so-called ``chart construction,'' in which each tree node is mapped to
an annotation graph arc.  An example tree and its corresponding annotation
graph are shown below:

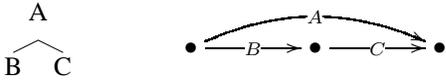
\begin{tabular}{ll}
\begin{minipage}{.25\linewidth}
\Tree [.A B C ]
\end{minipage}
&
\begin{minipage}{.7\linewidth}
\xymatrixcolsep{3pc}
\xymatrix{
\bullet \ar@/^1pc/[rr]|A \ar[r]|B &
\bullet \ar[r]|C &
\bullet
}
\end{minipage}
\end{tabular}

This approach has two shortcomings.  First, in the situation where a non-terminal
has a single child, the annotation graph is ambiguous.  Thus, the following two
simple trees have the same annotation graph representation:

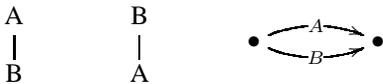
\begin{tabular}{lll}
\begin{minipage}{.15\linewidth}
\Tree [.A B ]
\end{minipage}
&
\begin{minipage}{.15\linewidth}
\Tree [.B A ]
\end{minipage}
&
\begin{minipage}{.6\linewidth}
\xymatrixcolsep{3pc}
\xymatrix{
\bullet \ar@/^.5pc/[r]|A \ar@/_.5pc/[r]|B &
\bullet
}
\end{minipage}
\end{tabular}

\noindent
The second shortcoming is that the annotation graph representation cannot express
discontinuous constituency (i.e. trees that contain crossing lines).

Both problems can be addressed by using equivalence classes or cross
references \cite{BirdLiberman01}.  We depict the relation between a child
arc and its parent using a dotted arrow, as shown below.  While this is partly
redundant, it involves minimal overhead.
\vspace{1ex}

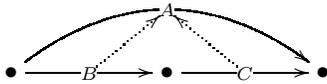
\xymatrixcolsep{4pc}
\xymatrix{
\bullet \ar@/^2pc/[rr]|A="A" \ar[r]|B="B" &
\bullet \ar[r]|C="C" &
\bullet \ar@{.>} "B";"A" \ar@{.>} "C";"A"
}
\vspace{1ex}

The elementary tree operations that we discussed above can now be implemented
directly in terms of the annotation graph model.  We begin with some
definitions.  Let $x$.start (resp.\ $x$.end) be the start (resp.\ end)
anchor of annotation $x$.  Let $\bar{x}$ be $x$'s
parent (undefined if $x$ has no parent).
Define $x$'s right sibling as follows:

\[
\overrightarrow{x} =
\left\{
\begin{array}{ll}
y & \mbox{if } y.\mbox{start} = x.\mbox{end}, \bar{x} = \bar{y} \\
\mbox{undefined} & \mbox{otherwise}
\end{array}
\right.
\]

Annotation graph arcs are typed, and our implementation requires two types,
namely ``word'' for word arcs (the orthographic string), and ``phrasal''
for the phrasal arcs.
Now we can define the above tree operations in terms of annotation graphs.

\begin{description}
\item[move down]
Given the arc $x$, insert a new coterminous arc which becomes
the parent of $x$.

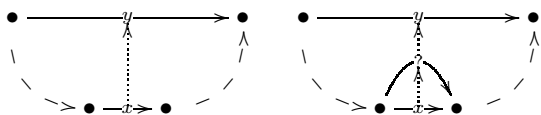
\xymatrixcolsep{1.5pc}
\begin{tabular}{ll}
\xymatrix{
\bullet \ar[rrr]|y="y"
  \ar@{-->}@/_1pc/[dr] &&& \bullet \\
& \bullet \ar[r]|x="x"
& \bullet \ar@{-->}@/_1pc/[ur] \\
\ar@{.>} "x";"y"
}
&
\xymatrix{
\bullet \ar[rrr]|y="y"
  \ar@{-->}@/_1pc/[dr] &&& \bullet \\
& \bullet \ar[r]|x="x" \ar@/^1.5pc/[r]|?="?"
& \bullet \ar@{-->}@/_1pc/[ur] \\
\ar@{.>} "x";"?"
\ar@{.>} "?";"y"
}
\end{tabular}

\item[promote right]
Move a rightmost child to the right, out of the subtree;
$x$'s parent ($y$) becomes $x$'s left sibling.  Note that
$y$ must be a phrasal arc.

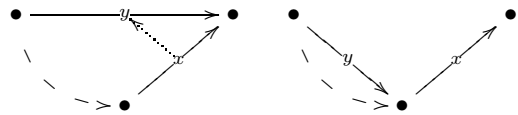
\xymatrixcolsep{2.5pc}
\begin{tabular}{ll}
\xymatrix{
\bullet \ar[rr]|y="y"
  \ar@{-->}@/_1pc/[dr] && \bullet \\
& \bullet \ar[ur]|x="x"
\ar@{.>} "x";"y"
}
&
\xymatrix{
\bullet \ar[dr]|y
  \ar@{-->}@/_1pc/[dr] && \bullet \\
& \bullet \ar[ur]|x
}
\end{tabular}

\item[demote right]
Move a subtree right, to become the leftmost daughter;
$x$'s right sibling $y$ becomes $x$'s parent.  Note that
$y$ must be a phrasal arc.

{\xymatrixcolsep{2.5pc}
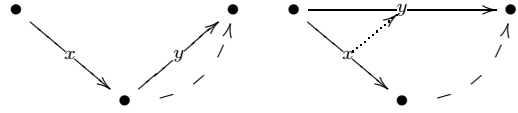
\begin{tabular}{ll}
\xymatrix{
\bullet \ar[dr]|x && \bullet \\
& \bullet \ar[ur]|y
  \ar@{-->}@/_1pc/[ur]
}
&
\xymatrix{
\bullet \ar[rr]|y="y"
  \ar[dr]|x="x" && \bullet \\
& \bullet \ar@{-->}@/_1pc/[ur]
\ar@{.>} "x";"y"
}
\end{tabular}
}
\end{description}

Observe that none of these operations alter the content or arrangement of
the word arcs.

\section{Dependency Treebanks}
\label{sec:dependency}

Dependency grammar is an approach to syntactic representation in which
words are organized into a hierarchy using a binary ``dependency''
relation.  Dependency trees pose a different set of challenges for
representation and manipulation, as discussed in this section.

\subsection{Survey}
\label{sec:depsurv}

The Turin University Treebank \cite{Bosco00} provides an example of a pure
dependency structure, showing a binary relation between the words.  The
treebank consists of 500 sentences, available from
\myurl{http://www.di.unito.it/~tutreeb/}.  A sample follows

\begin{sv}
1 E' (ESSERE VERB MAIN IND PRES INTRANS 3 SING) [0;TOP-VERB]
2 italiano (ITALIANO ADJ QUALIF M SING) [1;PREDCOMPL-SUBJ]
3 , (#\, PUNCT) [1;OPEN-PARENTHETICAL]
4 come (COME CONJ SUBORD MOD+TEMPO) [1;PREPMOD]
5 progetto (PROGETTO NOUN COMMON M SING) [4;PREPARG]
6 e (E CONJ COORD) [5;COORD]
7 realizzazione (REALIZZAZIONE NOUN COMMON F SING REALIZZARE
  TRANS) [6;COORD-2ND]
8 , (#\, PUNCT) [1;CLOSE-PARENTHETICAL]
9 il (IL ART DEF M SING) [1;SUBJ]
10 primo (PRIMO ADJ ORDIN M SING) [11;ADJCMOD-ORDIN]
11 porto (PORTO NOUN COMMON M SING) [9;NBAR]
12 turistico (TURISTICO ADJ QUALIF M SING) [11;ADJCMOD-QUALIF]
13 dell' (DI PREP MONO) [11;PREPMOD-LOC-SPEC]
13.1 dell' (LA ART DEF F SING) [13;PREPARG]
14 Albania (|Albania| NOUN PROPER) [13.1;NBAR]
\end{sv}

This format consists of:
the index of the word in the sentence;
the word;
parentheses containing the lemma and its morphosyntactic features;
brackets containing a reference to the parent of this dependent
and the name of the grammatical relation.

The Prague Dependency Treebank (PDT) \cite{Hajicova00} is a corpus with three
distinct layers of annotation -- morphological, analytic (syntactic), and
tectogrammatial.  We won't address the morphological annotation in order to
focus on more tree and treelike structures.  Both analytic and
tectogrammatical structures are represented as hybrid dependency trees, mixing
a pure dependency relation over the words with a minimum of constituents.
This representation is indicative of the underlying grammatical theory,
functional generative grammar. As the corpus uses an extensive tagset and
views annotations via a special tool, we refer the reader to the url above for
data samples.  PDT has an online tree viewer available
(see \myurl{http://shadow.ms.mff.cuni.cz/pdt/}).

The editor for the analytic level restricts the user to operations that
maintain a well formed dependency tree with constituent nodes mixed in.  In
accordance with the relatively free word order in Czech, the tool allows
movement of subtrees to arbitrary nodes, along with the creation and deletion
of constituents.

Further discussion of the tectogrammatical annotation is deferred to section
\S\ref{sec:semantic}

The TIGER Project uses a model intermediate between conventional trees and
dependency trees, represented in XML \cite{Mengel00}.  The dependency
structure is represented as a collection of nodes (\texttt{n} elements) and
words (\texttt{w} elements) connected using \texttt{edge}s.\footnote{ A
more abstract version of the same idea is described by \newcite{IdeRomary00}.}
A simplified version is shown below:

\begin{sv}
<n id="n1_500" cat="S">
  <edge href="#id(w1)"/>
  <edge href="#id(w2)"/>
</n>

<w id="w1" word="the"/>
<w id="w2" word="boy"/>
\end{sv}

This format can represent arbitrary digraphs.  The linear ordering of the
children of any given node is represented by the file order of the
corresponding elements (or by the internal structure of node identifiers).

An important property of this format is its extensibility.  For instance, edges can
be typed (with an attribute \texttt{type}, and coreference is marked
using edges having \texttt{type="semantic"}.  Edges can also be labeled
with the grammatical role of their dependent (e.g. \texttt{label="HD"} for
the head daughter).

\subsection{API}

An API for the structural editing of pure dependency trees is remarkably
simple.  We start with an arbitrary root node, and make all the words dependent
upon this node.  From this point, we can create any dependency relation by
iterative application of a single \textbf{move subtree} operation, which takes
a source node other than the root and a target node and makes one dependent
upon the other.  Thus, after an annotator identifies a single dependency, we
may see a tree as follows.

\begin{description}
\item[Tree 1]
\Tree [.Root w$_1$ w$_2$  [.w$_4$  w$_3$ ] ]
\end{description}

Since the word order is free, it may be that w$_1$ is dependant on w$_4$.  To
accomodate for this, we can either let the branches of the tree cross and
retain the terminal order, or we can rearrange the terminal order so that the
branches don't cross.  After \textbf{move subtree} is applied to source w$_1$
and target w$_4$, we would attain the following tree

\begin{description}
\item[Tree 2]
\Tree [.Root w$_2$ [.w$_4$ w$_1$ w$_3$ ] ]
\end{description}

But some systems may use an underlying grammar which mixes pure dependency
structure and a constituent based approach, as is found in the PDT.  Such an
approach allows the insertion of constituent nodes, equivalent to the
\textbf{move down} operation described for basic trees in \S\ref{sec:conapi}

\begin{description}
\item[Tree 3]
\Tree [.Root w$_1$ w$_2$ [.C w$_3$ ] w$_4$ ]
\end{description}

Such a constituent may then interact with the others just like the pure
dependency nodes associated with a single word.  For example, after two
\textbf{move subtree} operations, we may end up with the following.

\begin{description}
\item[Tree 4]
\Tree [.Root w$_2$ [.C [.w$_3$ w$_1$ ] w$_4$ ] ]
\end{description}

A user interface may facilitate a \textbf{delete} command which takes all the
children of a proper constituent node and moves them to the parent of the
deleted node, deleting the resulting empty constituent.

\begin{description}
\item[Tree 5]
\Tree [.Root w$_2$ [.w$_3$ w$_1$ ] w$_4$ ]
\end{description}

\subsection{AG Implementation}

To implement editable dependency trees with annotation graphs, we begin by
defining a root node as an arc which spans the length of the sentence.  As with
basic trees, each node in this tree has a parent pointer which by default
points to the root.  The primary editing operation is \textbf{move subtree},
which takes a tree and two distinguished nodes (w$_1$ and w$_2$), setting the
parent of w$_1$ to w$_2$.  This operation is sufficiently expressive to define
any structural editing operation on a pure dependency tree. 

Below we show a simple AG implementation of the editing operation \textbf{move
subtree} with source w$_1$ and target w$_4$.

\begin{description}
\item[Tree 1]

\xymatrixcolsep{3pc}
\xymatrix{
\bullet \ar[r]|*\txt{w1}="w1" \ar@/^4pc/[rrrr]|R="R" &
\bullet \ar[r]|*\txt{w2}="w2" &
\bullet \ar[r]|*\txt{w3}="w3" &
\bullet \ar[r]|*\txt{w4}="w4" &
\bullet
\ar@{.>} "w1";"R"
\ar@{.>} "w2";"R"
\ar@{.>}@/^1.5pc/ "w3";"w4"
\ar@{.>}@/^/ "w4";"R"
}

\item[Tree 2]

\xymatrixcolsep{3pc}
\xymatrix{
\bullet \ar[r]|*\txt{w1}="w1" \ar@/^4pc/[rrrr]|R="R" &
\bullet \ar[r]|*\txt{w2}="w2" &
\bullet \ar[r]|*\txt{w3}="w3" &
\bullet \ar[r]|*\txt{w4}="w4" &
\bullet
\ar@{.>}@/^3pc/ "w1";"w4"
\ar@{.>} "w2";"R"
\ar@{.>}@/^1.5pc/ "w3";"w4"
\ar@{.>}@/^/ "w4";"R"
}
\end{description}

For hybrid systems which allow constituents, we want to constrain the length
of the constituent arcs as much as possible.  In spite of the fact that
setting the length of these arcs to a constant would reduce overhead, we take
this approach in anticipation that the quasi-ordering over annotations will
provide a more substantial basis for layered annotation than following
pointers.  

We proceed by superimposing the implementation of \textbf{move up} and
\textbf{move down} directly on top of this and extend the definition of
\textbf{move subtree} so that it works on arbitrary constituents and maintains
a well formed hybrid structure.  We have developed an algorithm for this
which
requires the ability to distinguish between words and proper
constituents as well as between proper constituents and the root node.  We
accomplish this simply by checking the type of the arcs involved.  We
illustrate these extensions showing annotation graph representations of trees
$3$ and $4$ below.

\begin{description}
\item[Tree 3]
\xymatrixcolsep{3pc}
\xymatrix{
\bullet \ar[r]|*\txt{w1}="w1" \ar@/^4pc/[rrrr]|R="R" &
\bullet \ar[r]|*\txt{w2}="w2" &
\bullet \ar[r]|*\txt{w3}="w3" \ar@/^1.5pc/[r]|C="C" &
\bullet \ar[r]|*\txt{w4}="w4" &
\bullet
\ar@{.>} "w1";"R"
\ar@{.>} "w2";"R"
\ar@{.>} "w3";"C"
\ar@{.>} "w4";"R"
\ar@{.>} "C"; "R"
}

\item[Tree 4]

\xymatrixcolsep{3pc}
\xymatrix{
\bullet \ar[r]|*\txt{w1}="w1" \ar@/^4pc/[rrrr]|R="R" &
\bullet \ar[r]|*\txt{w2}="w2" &
\bullet \ar[r]|*\txt{w3}="w3" \ar@/^2pc/[rr]|C="C" & 
\bullet \ar[r]|*\txt{w4}="w4" &
\bullet
\ar@{.>}@/^2pc/ "w1";"w3"
\ar@{.>} "w2";"R"
\ar@{.>} "w3";"C"
\ar@{.>} "w4";"C"
\ar@{.>} "C"; "R"
}
\end{description}

\section{Treebanks and Semantic Trees}
\label{sec:semantic}

\subsection{Survey}
While many semantic relations are described in treebanks, predicate argument
structure remains the most commonly and systematically explored.  Each treebank
formulates some schema to represent the argument structure of clausal verbs,
and indeed this information is to some extent explicit in the parse itself.  To
complete the picture, the nodes of the parse tree are often decorated with
labels denoting more abstract relations.  In some cases, an
entire extra level of annotation is supplied separately in a parallel corpus,
as in the Prague Dependency Treebank (PDT).  In this section we
catalog a variety of predicate argument schemas, observing commonalities, and
exploring requirements inherent in capturing predicate argument structures with
treebanks.

The Susanne Corpus, developed as a by-product of a parsing schema for
unambiguous syntactic annotation, provides perspicuous coverage of predicate
argument structure of clausal verbs.  It decorates nodes with a variety of
function tags, though it restricts their usage to immediate constituents of
clauses.  

\begin{sv}
[Nns:s John] expected [Nns:O999 Mary] [Ti:o [s999 GHOST] 
to admit [Ni:o it]]
\end{sv}

The example above is similar to the Penn Treebank example in that it requires
coindexed nodes, but unlike the English Propbank, it does not use references to
syntactic nodes.  The complexity of predicate argument well-formedness
constraints together with a close coupling of syntactic and argument relations
are noteworthy by-products of embedding these relations in the syntactic schema.

We examine the tectogrammatical level of annotation in the PDT, as it
represents a more abstract linguistic structure closely related to 
predicate argument structure.  These trees are of the hybrid
dependency variety described in \S\ref{sec:dependency}  The tectogrammatical
dependency trees are roughly parallel to the analytic ones and their structure
is derived by deleting and adding nodes to the analytic trees.  
Spurious elements
of the surface string are removed and dropped arguments are added.
While these operations produce the structure of the tree, edge
labels such as {\em actor}, {\em patient}, {\em addressee}, {\em location}
denote semantic roles and modifiers.

The Penn Treebank uses attributes of phrase labels in
conjunction with grammatical relations to describe predicate argument
structure.  In the example below, the last nominal phrase is decorated with a
\texttt{LGS} tag denoting logical subject.  The syntactic
environment indicates the remaining parts of the argument structure, with the
head verb taking the role of the predicate and the preceding noun phrase taking
on the role of direct object.

\begin{sv}
(S
  (NP-SBJ (PRP they) )
  (VP (VBP attribute)
    (NP (-NONE- *T*-1) )
    (ADVP-MNR (RB directly) )
    (PP-CLR (TO to)
      (NP
        (NP (NNS forces) )
        (VP (VBN controlled)
          (NP (-NONE- *) )
          (PP (IN by)
              (NP-LGS (NNP PLO) (NNP Chairman) 
                  (NNP Yasser) (NNP Arafat))))))))
\end{sv}

Algorithms for extracting predicate argument structure, even from such
rich syntactic data, are faced with numerous complexities and ambiguities.  For
example, ghost constituents without explicit referents should be resolved,
disjoint constituents may form arguments, prepositional phrases may or may not
constitute arguments, and this information tends to be lexicalized over the
predicates \cite{PalmerRosenzweig01}.

As a next step, the English Propbank is under development, using the predicate
argument tagger mentioned above and hand-correcting the output.  The example of
this data below shows that the entire argument relation is explicitly marked.
Note that the argument label \texttt{ARG1} implicitly refers to specific syntactic
nodes rather than the surface string, in this case resolving the passive trace.
\begin{sv}
... they attribute  directly to forces controlled
 by PLO Chairman Yasser Arafat .

rel:        controlled
ARG1:      *trace* -> forces
ARG0-by:    PLO Chairman Yasser Arafat
\end{sv}

Additionally, the constituents of a particular argument may be disjoint
as the utterance argument of a sentence like

\begin{sv}
"I'm going home", John said, "so I can get some sleep".

rel:        said
ARG0:       John
ARG1:       [I'm going home] [so I can get some sleep]
\end{sv}

Phrasal predicates, such as {\em give up}, are almost never dominated by a
single node, and so are treated similarly.

Another source of variation occurs with conjunctions over more than one
argument.  For example, the sentence below yields two propositions.

\begin{sv}
John drove Mary to the store and Mike home

rel:       drove
Arg0:      John
Arg1:      Mary
Arg2-to:   the store

rel:       drove
Arg0:      John
Arg1:      Mike
Arg2:      home
\end{sv}

In the English propbank, we witness argument structure using {\em references
to} syntactic annotation, a one to many relation from arguments to constituents
(also vice versa), and the marking of sentence-local equivalences to
resolve grammatical motion.

In conclusion, capturing predicate argument structure is of definite interest
in the development of treebanks.  In all the cases examined, an extra level of
indirection from the syntactic structure is required.  The English Propbank
makes use of explicit references to syntactic constituents, the Susanne Corpus
employs highly structured decoration of nodes, inducing relations between the
nodes, and the PDT utilizes {\em differences} against the syntactic structure,
replacing analytic with semantic functions and recovering dropped arguments as
necessary.

\subsection{API}
\label{semanticapi}

%
%
%
%
%
%
%

As predicate argument structure has quite varied treatment, we'll look at both
argument structure as treated with the Penn Treebank and argument structure as
in the Prague Dependency Treebank.   However, we will restrict ourselves to
working with predicate argument data as derived from syntactic data rather than
as derived from scratch in order to best address the extant tagging efforts in
this domain.

In the case of the English Propbank, the operations are not editing operations
on trees {\em per se}, but operations on relations between constituents in a
given tree.  For each instance of a predicate in some parsed text, we can
characterize a proposition as a $4$-tuple consisting of the predicate, its
arguments predicate, its modifiers, and an equivalence relation over the nodes
in the parse tree.  Each of the arguments or modifiers consists of a label and
a non empty {\em set} of constituents, denoting its surface string content.
While this set of constituents is often singleton, any non-singleton set of
constituents represents a surface string which is not dominated by a single
node (this occurs with phrasal verbs and often with the utterance argument in
verbs of saying).  The equivalence relation over the nodes of the parse serve
to recover dropped arguments (as occurs with empty constituents) and
sentence-local antecedents of pronouns.  The case of conjunctions whose
conjuncts are not dominated by a single syntactic node is handled by
associating multiple propositions with the instance of the predicate (or lemma)
at hand.

The editing operations for the annotation process consist of associating
argument labels (e.g. arg0 \ldots argN) with constituents and identifying
equivalent nodes of the parse.  For example, annotating the argument structure
of the predicate {\em swim} on the parse tree below (with nodes identified in
terms of their leftmost terminal number and height) would yield a single
proposition whose predicate is $\{ (3,0) \}$, whose arguments consist of $\{
(Arg0, \{(2,0)\}) \}$, whose modifiers are $\emptyset$, and whose equivalences
are $\{((2,0), (0,0))\}$.

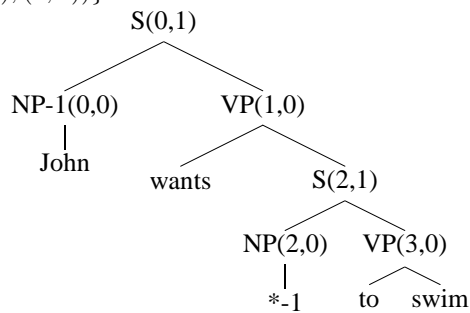
\Tree [.S(0,1) [.NP-1(0,0) John ] [.VP(1,0) wants [.S(2,1) [.NP(2,0) *-1 ] [.VP(3,0) to swim ] ] ] ]
\vspace{1ex}
In the PDT tectogrammatical annotation, the operations are structurally similar
to those of the analytic annotation, except that dropped arguments are added to
the structure and words can be deleted.  We defer addressing these issues
for future work.

\subsection{Semantic Implementation}
\label{sec:semimpl}
We describe an implementation of propbank annotation with annotation graphs.
Given an annotation graph parse of a basic tree as described in
\S\ref{sec:conimpl}, we first define the predicating lemma over a set of
constituents as an arc whose start point is the minimum of the start points of
the associated constituents and whose end point is the maximum of the end
points of the associated constituents.   For example, if the sentence is
\\
\begin{small}\texttt{
$\alpha_1$ John $\alpha_2$ belongs $\alpha_3$ to $\alpha_4$ the $\alpha_5$ club $\alpha_6$
}
\end{small}
\\
and $\alpha_n$ is an annotation graph anchor, and our predicating lemma is
{\em belongs to}, then the arc defining our predicate will start at $\alpha_2$
and end at $\alpha_4$.  Just as pointers were added for basic tree constituents,
we add sets of pointers to this arc to the constituents containing {\em belong}
and {\em to}.  This arc gets a label indicating that it is the predicating label,
say \texttt{pred}.

The arguments and modifiers of the lemma are denoted similarly, with an
appropriate label for the item in question.  The end-product is diagrammed
below:
\vspace{1ex}

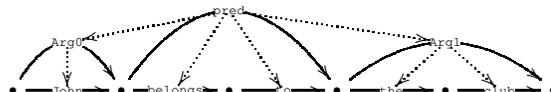
\texttt{
\tiny{
\xymatrixcolsep{2.6pc}
\xymatrix{
\bullet \ar[r]|*\txt{John}="John" \ar@/^1.5pc/[r]|*\txt{Arg0}="Arg0" &
\bullet \ar[r]|*\txt{belongs}="belongs" \ar@/^2.5pc/[rr]|*\txt{pred}="pred" &
\bullet \ar[r]|*\txt{to}="to"  & 
\bullet \ar[r]|*\txt{the}="the" \ar@/^1.5pc/[rr]|*\txt{Arg1}="Arg1" &
\bullet \ar[r]|*\txt{club}="club" &
\bullet
\ar@{.>} "Arg0";"John"
\ar@{.>} "pred";"belongs"
\ar@{.>} "pred";"to"
\ar@{.>} "Arg1";"the"
\ar@{.>} "Arg1";"club"
\ar@{.>} "pred";"Arg0"
\ar@{.>} "pred";"Arg1"
}
}
}
\vspace{1ex}

Finally, we specify the constituent equivalences by noting all the non
singleton equivalence classes whose members are among those associated
with a label.

\section{Discussion and Further Work}

Treebank formats and associated software tools are proliferating rapidly,
with little consideration for interoperability.  We have surveyed a wide
variety of treebank structures and operations, and shown how they can be
mapped onto the annotation graph model.  This has two important
ramifications, distinguishing our work from previous work.  First, the
false dichotomy between conventional trees and dependency trees goes away;
both types along with hybrid structures can be represented in a uniform
framework.  Second, a single comprehensive framework is used for both tree
and non-tree annotations, an integration that greatly facilitates
multilayer queries.

Several aspects of the survey and the analysis are incomplete, and we list
just three areas here.  First, there is another class of treebanks used for
grammar development, usually consisting of hand-crafted sentences
illustrating a particular linguistic phenomenon.  Each sentence is
associated with the correct analysis, expressed in a particular syntactic
formalism such as HPSG \cite{PollardSag94}.  An example of this kind of
corpus is the HPSG Treebank for Polish \cite{Marciniak00}.  Representing
such treebanks using annotation graphs would require a more expressive
model of arc labels than is currently permitted
(namely attribute-value matrices).

A second open question is in the area of bidirectionality.  Texts may
involve a mixture of directionality, such as an Arabic text containing
stretches of English.  In such texts, there is no longer a transparent
relationship between the sequence of orthographic words and their sequence
in a spoken utterance; the linguistic representation needs to encompass
both orderings somehow, even though annotation graphs force us to choose
one of the orderings as primary.

A third area for further investigation is query.  Now that the annotations
are all expressed in the same framework, how do we want to express queries
over the annotations?  A range of tree query languages have been proposed,
as discussed by \newcite{CassidyBird00}.  It is highly unlikely that a
single tree query language will ever meet the requirements of all research
projects.  Instead, we plan to investigate a number of tree query languages
and their mapping to a low-level annotation graph query language, such as
the one proposed by \newcite{BirdBunemanTan00}.

In this article we have surveyed treebanks, examining their data formats
and editing operations.  We have found that the existing treebank models do
not accomodate overlayed annotation very well.  We have developed abstract
APIs for treebanking operations which encompass the requirements of
conventional trees, dependency trees, and even predicate argument
structure.  We have described how these APIs may be directly implemented
using annotation graphs.  This facilitates multi-layered annotations and
leverages the array of annotation types that are already supported by
the annotation graph model.
\vspace{-6pt}

\section*{Acknowledgements}

This material is based upon work supported by the National Science
Foundation under Grant No.\ 9910603 (``International Standards in
Language Engineering'') and Grant Nos.\ 9978056, 9980009 (``Talkbank''),
and DARPA Grant No.\ MDA904-00C-2136.
\vspace{-6pt}

\small\raggedright
\bibliographystyle{lrec2000}

\end{document}